\documentclass{article}

\usepackage[final, nonatbib]{neurips_2024}

\usepackage[utf8]{inputenc} 
\usepackage[T1]{fontenc}    
\usepackage{url}            
\usepackage{booktabs}       
\usepackage{amsfonts}       
\usepackage{nicefrac}       
\usepackage{microtype}      
\usepackage{xcolor}         
\usepackage{amsmath}
\usepackage{extpfeil}
\usepackage{multirow}
\usepackage{makecell}

\definecolor{linkcolor}{RGB}{255,0,0}
\definecolor{urlcolor}{RGB}{255,105,180}
\definecolor{citecolor}{RGB}{66,168,235}
\usepackage[pagebackref=true,breaklinks=true,letterpaper=true,colorlinks,bookmarks=false]{hyperref}
\hypersetup{colorlinks=true,linkcolor=linkcolor,urlcolor=urlcolor,citecolor=blue}

\title{PCoTTA: Continual Test-Time Adaptation for \\ Multi-Task Point Cloud Understanding}

\author{%
  Jincen Jiang\thanks{Equal contributions.}\\
  Bournemouth University\\
  \texttt{jiangj@bournemouth.ac.uk} \\
  \And
  Qianyu Zhou\footnotemark[1]\\
  Shanghai Jiao Tong University\\
  \texttt{zhouqianyu@sjtu.edu.cn} \\
  \And
  Yuhang Li\\
  Shanghai University\\
  \texttt{yuhangli@shu.edu.cn}\\
  \And
  Xinkui Zhao\thanks{Corresponding authors.}\\
  Zhejiang University\\
  \texttt{zhaoxinkui@zju.edu.cn} \\
  \And
  Meili Wang\\
  Northwest A\&F University\\
  \texttt{wml@nwsuaf.edu.cn}\\
  \And
  Lizhuang Ma\\
  Shanghai Jiao Tong University\\
  \texttt{lzma@sjtu.edu.cn} \\
  \And
  Jian Chang\\
  Bournemouth University\\
  \texttt{jchang@bournemouth.ac.uk} \\
  \And
  Jian Jun Zhang\\
  Bournemouth University\\
  \texttt{jzhang@bournemouth.ac.uk} \\
  \And
  Xuequan Lu\footnotemark[2]\\
  La Trobe University\\
  \texttt{b.lu@latrobe.edu.au} \\
}

\begin{document}

\maketitle

\begin{abstract}
In this paper, we present PCoTTA, an innovative, pioneering framework for Continual Test-Time Adaptation (CoTTA) in multi-task point cloud understanding, enhancing the model's transferability towards the continually changing target domain. We introduce a multi-task setting for PCoTTA, which is practical and realistic,
handling multiple tasks within one unified model during the continual adaptation. Our PCoTTA involves three key components: automatic prototype mixture (APM), Gaussian Splatted feature shifting (GSFS), and contrastive prototype repulsion (CPR). Firstly, APM is designed to automatically mix the source prototypes with the learnable prototypes with a similarity balancing factor, avoiding catastrophic forgetting. Then, GSFS dynamically shifts the testing sample toward the source domain, mitigating error accumulation in an online manner. In addition, CPR is proposed to pull the nearest learnable prototype close to the testing feature and push it away from other prototypes, making each prototype distinguishable during the adaptation. 
Experimental comparisons lead to a new benchmark,  demonstrating PCoTTA's superiority in boosting the model's transferability towards the continually changing target domain. \emph{Our source code is available at}: https://github.com/Jinec98/PCoTTA.
\end{abstract}
\section{Introduction}
\label{sec:intro}
Recent advancements in 3D point cloud understanding have marked a significant leap in the field of computer vision \cite{jiang2024dhgcn, shao2024trici, xu2021paconv, thomas2019kpconv} and 3D processing \cite{de2023iterativepfn,de2024straightpcf, fung2024semreg, luo2021diffusion}.
Current methods \cite{qi2017pointnet, wang2019dynamic} primarily concentrate on training and testing on a single domain \cite{pang2022masked, jiang2023masked}. Nevertheless, they encounter noticeable performance drops on other target data. Different datasets have domain gaps, also known as domain shifts. For instance, models trained on meticulously structured synthetic data, such as ModelNet40 \cite{wu20153d}, may encounter difficulties in adapting to intricate and noisy real-world data, such as ScanObjectNN \cite{uy2019revisiting}. 

\begin{figure}[htb]
    \centering
    \includegraphics[width=1\linewidth]{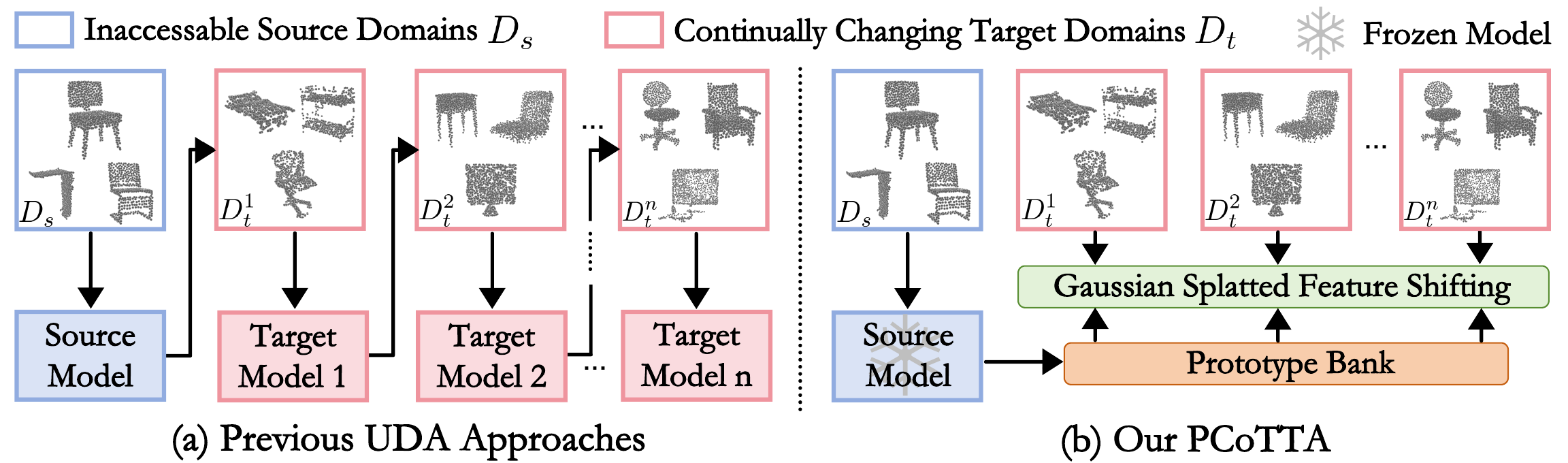}
    \caption{(a) Previous UDA approaches on point cloud suffer from catastrophic forgetting and error accumulation toward the continually changing target domains. (b) In contrast, we present an innovative framework PCoTTA to address these issues, enhancing the model's transferability. 
    }
    \label{fig:teaser}
\end{figure}

To mitigate domain shifts, recent researchers have introduced Unsupervised Domain Adaptation (UDA) techniques \cite{zhou2024test, long2024dgmamba, zhou2023self, zhou2023instance, zhou2023context} into point cloud understanding. Some studies synthesize diverse training data \cite{yi2021complete, wang2021cross, zhao2021epointda}, and others leverage adversarial learning~\cite{qin2019pointdan,zhang2021srdan,liu2021adversarial}, pseudo labeling~\cite{yang2021st3d,wu2023scoda,yang2022st3d++,hu2023density, shaban2023lidar}, consistency learning~\cite{wang2021cross,wang2023ssda3d,luo2021unsupervised,wei2022lidar}, feature disentanglement~\cite{jiang2021lidarnet} or self-supervised learning~\cite{achituve2021self,shen2022domain,zou2021geometry,liang2022point} to align the latent features across different domains. 
Nonetheless, these methods still face challenges especially when the target domain is streaming online and the whole training set of the target domain is inaccessible. As such, Test-Time Adaptation is introduced into point cloud~\cite{jiang2025dg,hatem2023point,hatem2023test} where the model can adapt to target distributions in an online manner at test-time without requiring any prior knowledge of the whole target domain. However, these methods may still fail when the target domain is continually changing, referred to as Continual Test-Time Adaptation (CoTTA), and such an open problem is rarely explored in point cloud understanding contexts. 

On the one hand, due to the lack of specific designs for 3D data,  current CoTTA methods~\cite{wang2022continual,dobler2023robust,sojka2023ar,niloy2024effective,brahma2023probabilistic,wang2024continual,gan2023decorate,niu2022efficient,yu2023noise} that are designed for 2D images are inapplicable to 3D point cloud tasks or exhibit less desired performance. On the other hand, few works like MM-CCTA~\cite{cao2023multi} target the CoTTA problem in 3D point cloud tasks. Although MM-CTTA~\cite{cao2023multi} designs a Continual Cross-Modal Adaptive Clustering (CoMAC) approach for 3D semantic segmentation, it suffers from two primary limitations: (1) it is specifically designed for one task only, and cannot handle other point cloud tasks such as point cloud reconstruction, denoising, and registration. Redesigning and retraining a CoTTA method for each task is cost-expensive. (2) The adapted model would inevitably forget the previously learned data (catastrophic forgetting) and accumulate the model errors (error accumulation) during the continual adaptation, limiting the model's transferability toward the target domains. 

Motivated by the above analysis, we present PCoTTA, an innovative, pioneering framework for Continual Test-Time Adaptation (CoTTA) in multi-task point cloud understanding, enhancing the model's transferability towards the continually changing target domain. 
Also, we introduce a multi-task setting for PCoTTA, which is practical and realistic,
handling multiple tasks within one unified model during the adaptation.
In particular, given an off-the-shelf model pre-trained on the source domains, our PCoTTA 
aims to bridge the gap between the source and continually changing target domains by dynamically scheduling the shifting amplitude at test time.

Our PCoTTA mainly consists of three novel modules. 
Firstly, to prevent catastrophic forgetting, we propose an automatic prototype mixture (APM) strategy that automatically mixes the source prototypes with the learnable target prototypes based on the automatic similarity balancing factor (ASBF), which avoids straying too far from its original source model. 
Secondly, to mitigate error accumulation, we present Gaussian Splatted feature shifting (GSFS) that dynamically shifts the testing sample toward the source domain based on the distance between the testing features and the shared prototype bank. 
In addition, we also introduce Gaussian weighted graph attention to further adaptively schedule the shifting amplitude in a learnable manner at test time. Our insight is to highlight the similarity between the target sample and its similar prototypes and suppress the dissimilar weights. 
It therefore mitigates the risk of catastrophic forgetting. Finally, we devise the contrastive prototype repulsion (CPR) to pull the nearest learnable prototype close to the testing feature and push it away from other prototypes, making learnable prototypes more distinguishable. 
Furthermore, we present a new benchmark. We meticulously select a total of $30,954$ point cloud samples from $4$  datasets, including $2$ synthetic datasets (ModelNet40 \cite{wu20153d} and ShapeNet \cite{chang2015shapenet}) and $2$ real-world datasets (ScanNet \cite{dai2017scannet} and ScanObjectNN \cite{uy2019revisiting}), encompassing $7$ same object categories, and generate corresponding ground truth for $3$ different tasks (reconstruction, denoising,
and registration). 
Our main contributions are three-fold: 
\begin{itemize}
    \item We present PCoTTA, an innovative, pioneering, and unified framework for Continual Test-Time Adaptation (CoTTA) in multi-task point cloud understanding, enhancing the model's transferability towards the continually changing target domain. We introduce a multi-task setting with a new benchmark for PCoTTA, which is practical and realistic in the real world.

    \item We devise three innovative modules for PCoTTA, \emph{i.e.,} automatic prototype mixture (APM), Gaussian Splatted feature shifting (GSFS), and contrastive prototype repulsion (CPR) strategies, where APM avoids straying too far from its original source model, mitigating the risk of catastrophic forgetting, and GSFS dynamically shifts the testing sample toward the source model, alleviating error accumulation, and CPR pulls the nearest learnable prototype close to the testing feature and pushes it away from other prototypes.
        
    \item Extensive experimental results with analysis demonstrate the effectiveness and superiority of our presented method, surpassing the state-of-the-art approaches by a large margin.
\end{itemize}

\section{Related Work}
\label{sec:related_work}
\noindent\textbf{Point Cloud Understanding.} Pioneering works such as PointNet~\cite{qi2017pointnet} and PointNet++~\cite{qi2017pointnet++} process point clouds directly, with PointNet~\cite{qi2017pointnet} utilizing pooling operations for spatial encodings and PointNet++~\cite{qi2017pointnet++} employing hierarchical processing for capturing local structures at various scales. DGCNN~\cite{wang2019dynamic} updates the graph in feature space to capture dynamic local semantic features, while PCT~\cite{guo2021pct} addresses global context and dependencies within point clouds using order-invariant attention mechanisms. Recent methods like Point-BERT~\cite{yu2022point} and Point-MAE~\cite{pang2022masked} have introduced Masked Point Modeling (MPM) for reconstructing obscured point clouds. Point-BERT~\cite{yu2022point} employs a BERT-style pre-training strategy for improving performance in subsequent tasks, while Point-MAE~\cite{pang2022masked} uses masked autoencoders for self-supervised learning, enabling comprehensive representations without labeled data. 
PIC~\cite{fang2024explore} explores the In-Context Learning (ICL) paradigm to enhance 3D point cloud understanding, showcasing the model's potential in multi-task learning.
Despite their gratifying progress, they only consider a single data domain and suffer from performance degradation in target domains. Thus, we study continual test-time adaptation for point cloud tasks.

\noindent\textbf{Continual Test-Time Adaptation.} This task aims to adapt the pre-trained model toward the continually changing environments at test time.  CoTTA~\cite{wang2022continual} employs a weighted augmentation-averaged mean teacher framework to address this issue. \cite{gong2022note} capitalizes on the temporal correlations within streamed input data through reservoir sampling and instance-aware batch normalization. \cite{gan2023decorate,yang2024exploring} introduce domain-specific prompts and domain-agnostic prompts to preserve both domain-specific and domain-shared knowledge, respectively. Meanwhile, EATA~\cite{niu2022efficient} focuses on adapting non-redundant samples to facilitate efficient updates.
Another work RMT~\cite{dobler2023robust} uses a mean teacher setup with symmetric cross-entropy and contrastive learning. 
More recently,  MM-CTTA~\cite{cao2023multi} designs a Continual Cross-Modal Adaptive Clustering (CoMAC) approach for 3D semantic segmentation.
Despite these methods showing promising potential in 3D data, they mainly suffer from two limitations: Firstly, they are specifically designed for one task only, and they cannot handle other point cloud tasks like those in PIC \cite{fang2024explore}. Secondly, the model would inevitably forget the previously learned knowledge (catastrophic forgetting) and accumulate prediction errors (error accumulation) during the continual adaptation, leading to undesirable results.  In contrast, we present a unified model,  PCoTTA, for continual test-time adaptation of multi-task point cloud understanding. 
\section{Method}
\label{sec:method}
\begin{figure}[htb]
    \centering
    \includegraphics[width=1\linewidth]{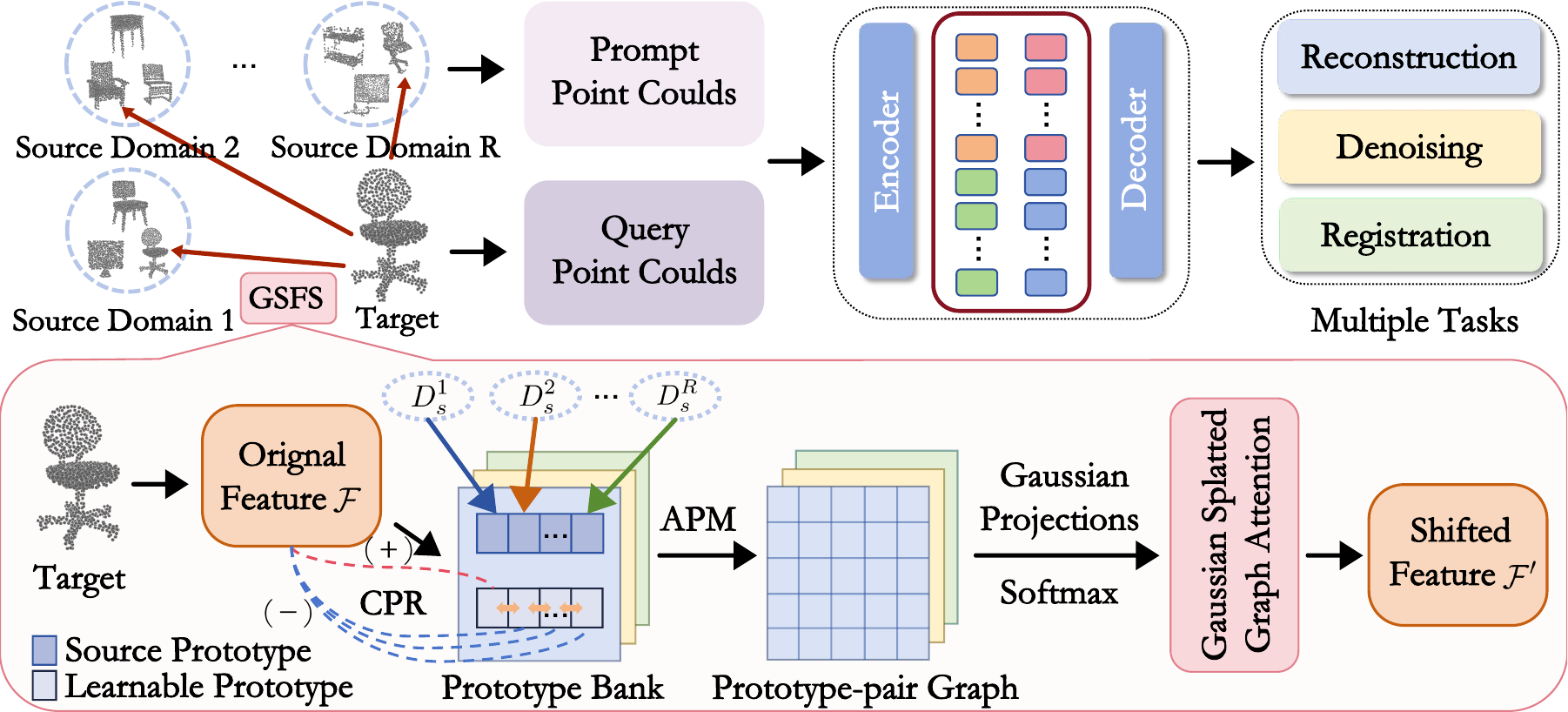}
    \caption{
    Our PCoTTA. It addresses continually changing targets by using their nearest source sample as a prompt for multi-task learning within a unified model. We introduce Gaussian Splatted Feature Shifting (GSFS) to align unknown targets with sources, improving transferability. Source prototypes from different domains and learnable prototypes form a prototype bank. The Automatic Prototype Mixture (APM) pairs these prototypes based on the similarity to the target, preventing catastrophic forgetting. We project these prototypes as Gaussian distributions onto the feature plane, with larger weights assigned to more relevant ones. Our graph attention updates these weights dynamically to mitigate error accumulation. Additionally, our Contrastive Prototype Repulsion (CPR) ensures that learnable prototypes are distinguishable for different targets, enhancing adaptability. 
    }
    \label{fig:overview}
\end{figure}

We present a novel framework, namely PCoTTA, for Continual Test-Time Adaptation in point cloud understanding tasks with the practical multi-task and multi-domain setting. As depicted in Figure \ref{fig:overview}, we propose an innovative approach to effectively address the challenges of continuously changing target data in test time within a unified model. In particular, our PCoTTA consists of three novel components: Automatic Prototype Mixture (APM) to mitigate catastrophic forgetting, Gaussian Splatted Feature Shifting (GSFS) to alleviate error accumulation, and Contrastive Prototype Repulsion (CPR) to make learnable prototypes distinctive across continually changing target domains. 

\subsection{Point Cloud Continual Test-Time Adaptation}

\noindent\textbf{Problem Formulation.}
In this work, we study a practical setting of continual test-time adaptation for multi-task point cloud understanding.
Suppose we have $R$ source domains $D_s=\{D_s^1, D_s^2, \dots, D_s^R\}$, our PCoTTA employs the input point clouds $\{I_q, I_p\}$ (along with their targets $\{T_q^k, T_p^k\}$, where $k$ represents the task index) from two different sources $\{D_s^i, D_s^j\} \in D_s, (i \neq j)$ to form the context pairs, facilitating the model with a comprehensive representation that effectively generalizes across all source domains. In the pre-training phase, each input sample comprises two context pairs: the input point cloud pair (query and prompt) and their corresponding target pair addressing the same task. 
During the test time, our PCoTTA strives to align streamed target data $I_t \in D_t$ (where $D_t = \{D_t^1 \cup D_t^2 \cup \dots\}$ denotes the set of continuously varying target domains) towards sources that possess correlative features to the off-the-shelf pre-trained model. 

\noindent\textbf{Multi-task Learning Objective.}
We follow PIC~\cite{fang2024explore} for three point cloud understanding tasks: (1) Reconstruction, which focuses on generating a dense point cloud from the sparse input; (2) Denoising, aiming at eliminating noise or outliers from the input point cloud; (3) Registration, dedicated to restoring the original orientation of a randomly rotated point cloud. Please note these three tasks might be slightly different from conventional definitions.
They are used as they can be handled similarly given current point learning can predict point positions directly. This makes them `unified' with position output and a single loss. 
We employ the MPM framework to generate query results across multiple downstream tasks, with a unified objective and a unified model. 
Let $\Phi(\cdot)$ denote the model shared across all domains and all tasks, and predicted masked patches $P$ can be depicted as:
\begin{equation}
    \{I_q, I_p\} \rightarrow P = \Phi(\mathcal{F}(I_q) \oplus \mathcal{F}(T_q^k) \oplus \mathcal{F}(I_p) \oplus \mathcal{F}(T_p^k), \mathcal{M}),
\end{equation}
where $\mathcal{F}(\cdot)$ represents the feature encoder that produces patch-wise features, \textit{i.e.,} the tokens, from point cloud feature, and $\mathcal{M}$ denotes the masked token utilized to replace the masked patches in the inputs. During the pre-training stage, $\mathcal{M}$ is derived from the random masking among query and prompt point clouds; whereas at test time, $\mathcal{M}$ exclusively masks the query target to generate the task-specific query output. The Chamfer Distance (CD) is used as the loss, measuring the similarity between the predicted masked patch $P$ and its corresponding ground truth $G$:
\begin{equation}
    \mathcal{L}_{cd} = \frac{1}{|P|} \sum_{x \in P} \min _{y \in G}\|x-y\|_2^2 + \frac{1}{|G|} \sum_{y \in G} \min _{x \in P}\|y-x\|_2^2.
\end{equation}

\subsection{Automatic Prototype Mixture}
The empirical evidence perceived by the human visual system illustrates that when people are not certain about the identity of an object, they would seek to find a distinct object from other domains that share high semantic similarity with the current object in the target domain. Motivated by this, we propose Automatic Prototype Mixture (APM) that adapts to continuously changing target data by aligning it with model-familiarized prototypes of source domains at test time.

\noindent \textbf{Source Prototypes Estimation.}
Our insight lies in that source prototypes can potentially represent source domains' feature distribution. Pulling the target data toward source prototypes within the feature space can effectively narrow the domain gap, bolstering the model's transferability.
Accordingly, the source prototypes $Z_s^i(i \in [1, R])$ can be determined by computing the average of all tokens produced by the MPM framework across all data within the sources:
\begin{equation}
    Z_s^i = \frac{1}{N_{D_s^i}} \sum_{n=1}^{N_{D_s^i}} \mathcal{F}(I_n), \quad Z_s \in \mathbb{R}^{R\times K \times M\times C},
\end{equation}
where $N_{D_s^i}$ denotes the sample number in domain $D_s^i$,  $K$ represents the tasks number, and $M$ indicates the tokens number in each sample. 
After pre-training on the multi-task and multi-domain setting, we save all source prototypes $Z_s$  derived from the model at the last epoch, considering them as the shared common knowledge available to the target data during the test time. 

\noindent\textbf{Prototype Bank.} 
We propose a novel prototype bank that stores not only the source prototypes $Z_s$ but also a series of \textit{learnable prototypes} $Z_l \in \mathbb{R}^{S\times K\times M\times C}$, where $S$ indicates the number of all potential target domains $D_t$. 
The learnable prototypes $Z_l$ aim to extract the current domain knowledge, thereby paving the way for handling subsequent unknown test data. 
We achieve the test-time adaptation of target tokens through the mixture of the paired prototypes in the prototype bank, selectively updating only the learnable prototypes while maintaining the source ones, thus mitigating the risk of catastrophic forgetting of the source domain knowledge due to the over-reliance on the adaptively learned information.

\begin{figure}[t]
    \centering
    \includegraphics[width=1\linewidth]{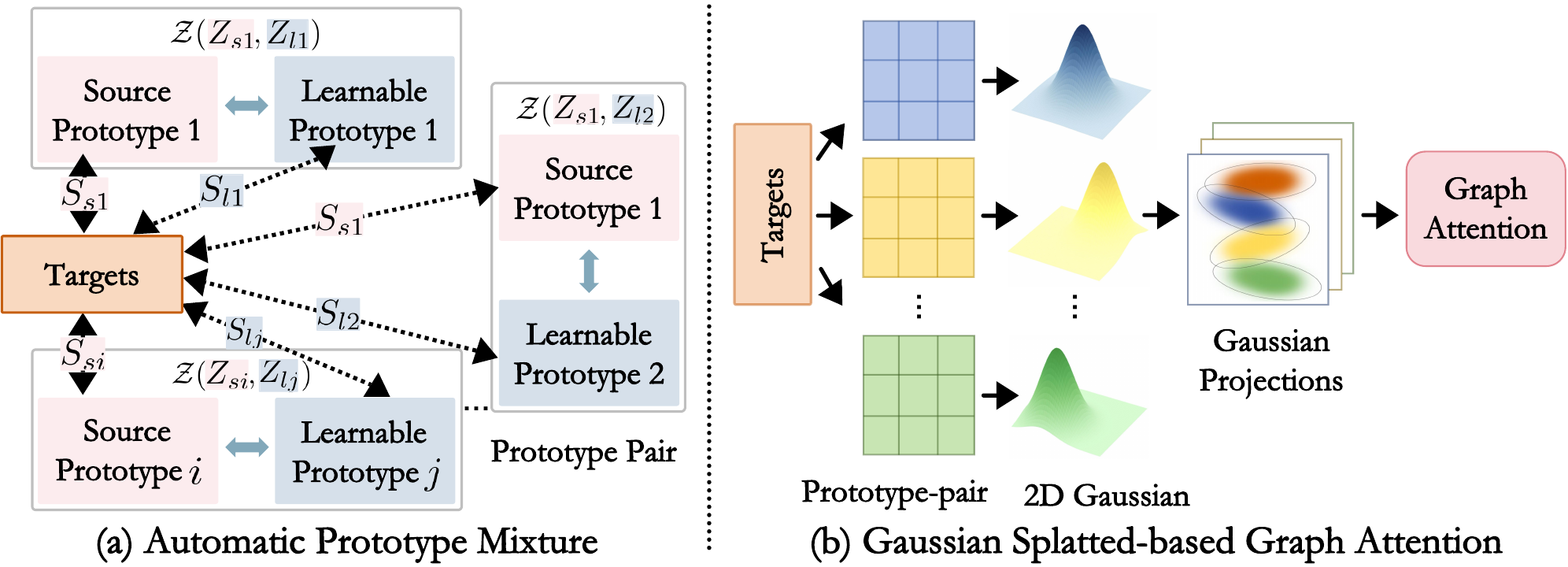}
    \caption{(a) Automatic Prototype Mixture (APM) considers both source and learnable prototypes with their similarities to the target, mitigating catastrophic forgetting by preserving source information. (b) Gaussian Spaltted-based Graph Attention enables dynamic updating weights among all prototype-pair nodes based on the Gaussian projections splatted onto the feature plane. }
    \label{fig:apm_gsfs}
\end{figure}

\noindent\textbf{Prototype-pair Node Mixture.}
The source prototypes $Z_s$ along with the learnable prototypes $Z_l$ in the prototype bank are paired to form prototype-pair nodes.
As illustrated in Figure \ref{fig:apm_gsfs}(a), the tokens from each test data $\mathcal{F}(I_t)$ serve as the central node in a graph structure, adjacent to all prototype-pair nodes.
We propose the Automatic Prototype Mixture (APM) module, designed to merge source and learnable prototypes within each node by considering their token-wise feature distances with the test data, \textit{i.e.,} the dot product between two feature vectors. 

Firstly, we need to repeat the test data tokens $\mathcal{F}(I_t)$ to align with the total number of prototypes:
\begin{equation}
    \mathcal{R}(I_t) = [ \underbrace{\mathcal{F}(I_t) \text{ } \mathcal{F}(I_t) \dots \mathcal{F}(I_t)}_{\text{repeat }x \text{ times}}],
\end{equation}
where $x$ equals $R$ or $S$. Then, the similarity $\mathcal{S}_s$ between source prototypes $Z_s$ and the test data $I_t$ is:
\begin{equation}
    \mathcal{S}_s = \frac{1}{M}\sum Diag(Norm(Z_s) \cdot Norm(\mathcal{R}(I_t)^T)) \in \mathbb{R}^{R \times K},
\end{equation}
where $Diag(\cdot)$ indicates creating a diagonal matrix, $Norm(\cdot)$ denotes normalization along the last dimension (\textit{i.e.,} the feature channel), and $(\cdot)^T$ represents transposition specifically applied to the last two dimensions. Likewise, the similarity $\mathcal{S}_l \in \mathbb{R}^{S \times K}$ between the test data and the learnable prototypes can also be determined. 

We further propose the Automatic Similarity Balancing Factor (ASBF) to measure the impact of the source and learnable prototypes toward the test data through the similarities $\mathcal{S}_s$ and $\mathcal{S}_l$, automatically prioritizing the prototypes and assigning greater weight to more similar components. The mixed prototypes (\textit{i.e.,} the prototype-pair nodes) $Z_m$ can be defined as:
\begin{equation}
    Z_m = \mathcal{Z}(Z_s, Z_l) = \frac{\mathcal{S}_s}{\mathcal{S}_s + \mathcal{S}_l} \cdot Z_s + \frac{\mathcal{S}_l}{\mathcal{S}_s + \mathcal{S}_l} \cdot Z_l \in \mathbb{R}^{R \times S \times K}.
\end{equation}

APM effectively considers the two types of prototypes while ensuring that the engagement with the original pre-trained model and source prototype is maintained, preventing catastrophic forgetting.

\subsection{Gaussian Splatted Feature Shifting}
Our PCoTTA considers all nodes but applies dynamically updated weights to each edge, enabling distinguishing the feature shifting in the continual test-time adaptation.
To this end, we propose the Gaussian Splatted Feature Shifting (GSFS), preventing error accumulation in an online
manner.

\noindent\textbf{Gaussian Splatted-based Graph Attention. }
Our key insight is that prototypes within a node (\textit{i.e.,} the source-learnable prototypes pair) can mutually constrain each other and collaboratively determine the weight of the edge connected to this node. 
We interpret the similarities between the test data and these two types of prototypes as Gaussian projections onto a plane, with the source and learnable prototypes corresponding to two orthogonal axes, respectively.
As illustrated in Figure \ref{fig:apm_gsfs}(b), the projections of all nodes on the feature plane are treated as a blend of Gaussians, where nodes with stronger correlations to the test data (\textit{i.e.,} higher similarities) are assigned larger weights. In this manner, all prototype-pair nodes are seamlessly integrated into the feature adaptation process. 
We compute the attention coefficient of each node as follows:
\begin{equation}
    \mathcal{E}(\mathcal{S}_s, \mathcal{S}_l) = \omega - \mathcal{G}(\mathcal{S}_s, \mathcal{S}_l)
    = \omega -  \frac{1}{2\pi\sigma_{\mathcal{S}_s}\sigma_{\mathcal{S}_l}} e^{-\frac{1}{2}\left(\frac{(\mathcal{S}_s - \mu_{\mathcal{S}_s})^2}{\sigma_{\mathcal{S}_s}^2} + \frac{(\mathcal{S}_l - \mu_{\mathcal{S}_l})^2}{\sigma_{\mathcal{S}_l}^2}\right)},
\end{equation}
where $\sigma_{\mathcal{S}_s}, \sigma_{\mathcal{S}_l}$ represent the variances of $\mathcal{S}_s$ and $\mathcal{S}_l$,  respectively, and $\mu_{\mathcal{S}_s}, \mu_{\mathcal{S}_l}$ denote their mean values. Note that the Gaussian function is inversely correlated with the similarity. We introduce a parameter $\omega$, set slightly above the maximum similarity observed, to ensure that more similar prototypes have a stronger influence. 

\noindent\textbf{Attention-based Feature Shifting.}
The attention coefficient $\mathcal{E}^{i,j}$ reflects the relative importance of the source $Z_s^i$ and the learned $Z_l^j$ prototypes. To ensure comparability across all connected nodes, we normalize coefficients using the Softmax function. Furthermore, we adopt a learnable shared attention module to dynamically update edge weights as follows:
\begin{equation}
    \mathcal{W}_{i,j} = \Psi_{\theta}(Softmax(\mathcal{E}^{i,j}))
    = \Psi_{\theta}(\frac{e^{\mathcal{E}^{i,j}}}{\sum_{m=1}^{N_{Z_m}} e^{\mathcal{E}^{i,m}}}),
\end{equation}
where $\Psi_{\theta}$ denotes a series of Convolution Layers parameterized by $\theta$, and $N_{Z_m}$ indicates the total number of the mixed prototypes (equals $R\times S$), indexed by $m$. 
Thereby, we can merge all prototype-pair nodes with the central node, \textit{i.e.,} the test data features, using the adaptive edge weights:
\begin{equation}
    \mathcal{F}'(I_t) = \frac{1}{R\times S} \sum_{i,j=0}^{R,S}((1-\mathcal{W}_{i,j}) \cdot \mathcal{F}(I_t) + \mathcal{W}_{i,j} \cdot Z_m^{i,j}).
\end{equation}

The proposed GSFS dynamically updates the contribution from each node in the graph with Gaussian Splatted-based graph attention, effectively assigning distinctive weights of feature shifting according to each node's relevance to the test data. This enables the test data to effectively align with task-beneficial domains, significantly diminishing the potential for error accumulation in the model. 

\subsection{Contrastive Prototype Repulsion}
The learnable prototypes within the prototype bank strive to capture the domain-specific knowledge of the current test data. Instead of predicting domain pseudo-labels to all test data, a common practice in prior techniques \cite{wang2022continual, gan2023decorate}, our method pulls the most similar learnable prototype closer to the test data while pushing it away from the others, thereby implicitly learning the distinctive features from different samples. To this end, we introduce Contrastive Prototype Repulsion (CPR) that effectively refines the learnable prototypes in the prototype bank, ensuring their distinctiveness and preventing domain-flattening from iterative learning and settling at sub-optimal points. 
We form a positive pair between the test data $\mathcal{F}'(I_t)$ and their nearest learnable prototype $Z_l^t$, and the rest serve as negative pairs. Our CPR optimization objective can be expressed as:
\begin{equation}
    \mathcal{L}_{pr} = - \frac{1}{S} \sum_{Z_l^{\cdot}\in S} \log\left(\frac{e^{{\mathcal{F}'(I_t) \cdot Z_l^t}/{\tau} }}{e^{{\mathcal{F}'(I_t) \cdot Z_l^t}/{\tau}} + \sum_{k \neq t} e^{{\mathcal{F}'(I_t) \cdot Z_l^k}/{\tau}}}\right),
\end{equation}
where $\tau$ is the temperature parameter, set to $0.07$ by default. 
Therefore, the overall loss function of our PCoTTA in the test time adaptation can be defined as follows:
\begin{equation}
    \mathcal{L} = \mathcal{L}_{cd} + \alpha \cdot \mathcal{L}_{pr},
\end{equation}
where $\alpha$ is the weighting factor that balances the two loss terms.

\section{Experiments}
\label{sec:exp}

\subsection{Experimental Setting}
\label{sec:experimental_setting}
\noindent\textbf{Implementation Details.}
We implement our method using PyTorch and perform experiments on two NVIDIA A40 GPUs. Following PIC \cite{fang2024explore}, we set the training batch size to $128$ and utilize the AdamW optimizer \cite{loshchilov2019decoupled}. The learning rate is set to $0.001$, with a cosine learning scheduler and a weight decay of $0.05$. All models are trained for $300$ epochs during the pertaining stage, and we train the pre-trained model for $3$ epochs on the source domains to initialize our prototype bank. 
At testing time, we continuously adapt test samples to the source pre-trained model and validate the anti-forgetting capability of our method across multiple rounds. 
Each point cloud is sampled to $1,024$ points and then split into $64$ patches, with each patch consisting of $32$ points. 
Within the MPM framework, the mask ratio is set to $0.7$, consistent with prior studies \cite{yu2022point, pang2022masked}.

\noindent\textbf{New Benchmark.}
We meticulously curate and select data from $4$ distinct datasets ($2$ synthetic and $2$ real-world datasets), containing $7$ identical object categories. 
Subsequently, we generate corresponding ground truth based on $3$ different tasks. 
The synthetic datasets include ModelNet40 \cite{wu20153d} and ShapeNet \cite{chang2015shapenet}. ModelNet40 consists of $3,713$ samples for training and $686$ for testing, while ShapeNet comprises $15,001$ training samples and $2,145$ testing samples. 
We also consider real-world data: ScanNet \cite{dai2017scannet} and ScanObjectNN \cite{uy2019revisiting}. ScanNet provides annotations for individual objects in real 3D scans, and we choose $5,763$ samples for training and $1,677$ for testing. ScanObjectNN includes $1,577$ training samples and $392$ testing samples. 
In all experiments, we employ ScanNet~\cite{dai2017scannet} and ShapeNet~\cite{chang2015shapenet} as the source domains and evaluate the transferability of our method on the other two target domains, \textit{i.e.,} ModelNet40~\cite{wu20153d} and ScanObjectNN~\cite{uy2019revisiting} with 3 repeated times by default.

\setlength{\tabcolsep}{1pt}
\begin{table}[htbp]
    \caption{Comparisons with the state-of-the-art approaches on the CoTTA setting. We report the Chamfer Distance (CD, $\times 10^{-3}$) for different tasks. The lower CD denotes the better performance.}
    \centering
    \resizebox{\textwidth}{!}{
    \begin{tabular}{l|c|ccc|ccc|ccc|ccc|ccc|ccc}
    \hline
         \multicolumn{2}{c|}{Time} &  \multicolumn{18}{c}{t\(\xlongrightarrow{\hspace{7cm}}\)}\\
    \hline
         \multicolumn{2}{c|}{Rounds} & \multicolumn{6}{c|}{1} & \multicolumn{6}{c|}{2} & \multicolumn{6}{c}{3} \\
    \hline
         \multicolumn{2}{c|}{Target Domains} & \multicolumn{3}{c|}{ModelNet40} & \multicolumn{3}{c|}{ScanObjectNN} & \multicolumn{3}{c|}{ModelNet40} & \multicolumn{3}{c|}{ScanObjectNN} &\multicolumn{3}{c|}{ModelNet40} & \multicolumn{3}{c}{ScanObjectNN}\\
    \hline
         Methods & Setting & Rec. & Den. & Reg.& Rec. & Den. & Reg.& Rec. & Den. & Reg.& Rec. & Den. & Reg.& Rec. & Den. & Reg.& Rec. & Den. & Reg.\\
    \hline
         PointNet \cite{qi2017pointnet} & \multirow{5}{*}{\makecell{Task-\\specific\\Models}} & 38.2 & 38.1 & 40.4 & 39.3 & 39.5 & 41.5 & 37.7 & 38.4 & 40.7 & 39.0 & 39.8 & 42.0 & 38.2 & 38.1 & 40.9 & 39.2 & 39.5 & 42.2\\
         DGCNN \cite{wang2019dynamic} & ~ & 36.0 & 33.7 & 36.0 & 37.3 & 35.6 & 37.6 & 35.3 & 32.7 & 34.1 & 36.6 & 34.6 & 36.0 & 36.1 & 32.6 & 34.7 & 37.1 & 34.4 & 36.5\\
         PCT \cite{guo2021pct} & ~ & 29.7 & 29.6 & 30.6 & 30.2 & 30.3 & 31.5 & 29.6 & 29.8 & 30.6 & 30.2 & 30.5 & 31.8 & 30.8 & 29.5 & 30.8 & 31.5 & 30.1 & 31.8\\
         Pointmixup \cite{chen2020pointmixup} & ~ & 37.3 & 36.8 & 38.5 & 38.4 & 37.0 & 40.3 & 37.0 & 36.5 & 37.9 & 38.9 & 36.7 & 40.1 & 37.8 & 36.8 & 38.1 & 38.5 & 36.9 & 40.7\\
         PointCutMix \cite{zhang2022pointcutmix} & ~ & 41.5 & 40.1 & 38.2 & 43.3 & 44.7 & 40.5 & 41.1 & 40.4 & 38.5 & 42.9 & 44.1 & 40.7 & 40.8 & 40.0 & 39.2 & 43.1 & 44.5 & 40.2\\
    \hline
         PointNet \cite{qi2017pointnet} &\multirow{5}{*}{\makecell{Multi-\\task\\Models}}& 38.3 & 38.8 & 41.4 & 39.5 & 40.4 & 43.0 & 38.0 & 38.5 & 41.3 & 39.3 & 40.2 & 42.8 & 38.4 & 38.6 & 42.1 & 39.6 & 40.4 & 43.3\\
         DGCNN \cite{wang2019dynamic} &~ & 37.0 & 33.5 & 36.0 & 38.1 & 35.2 & 37.7 & 36.9 & 33.2 & 36.0 & 38.1 & 35.5 & 37.7 & 36.9 & 33.2 & 36.5 & 38.0 & 35.2 & 37.8\\
         PCT \cite{guo2021pct} &~ & 29.6 & 30.2 & 32.5 & 30.4 & 30.9 & 33.7 & 29.9 & 30.4 & 32.4 & 30.7 & 30.9 & 33.5 & 29.8 & 30.0 & 31.9 & 30.5 & 30.8 & 33.1\\
         Pointmixup \cite{chen2020pointmixup} &~ & 37.8 & 41.5 & 39.2 & 44.6 & 45.1 & 40.7 & 38.3 & 40.9 & 39.1 & 43.4 & 44.2 & 41.6 & 38.2 & 41.3 & 39.2 & 44.1 & 44.8 & 40.9\\
         PointCutMix \cite{zhang2022pointcutmix} &~ & 42.3 & 44.1 & 39.9 & 45.4 & 47.3 & 43.8 & 41.9 & 43.2 & 40.1 & 45.2 & 46.8 & 42.3 & 42.1 & 43.7 & 40.4 & 45.2 & 47.1 & 42.9\\
    \hline
         Baseline \cite{fang2024explore} &\multirow{2}{*}{\makecell{ICL\\Models}}& 79.7 & 126.3 & 106.3 & 82.3 & 129.5 & 113.4 & 86.5 & 127.7 & 106.8 & 83.0  & 124.7 & 110.2 & 78.9 & 123.6 & 110.6 & 84.5 & 125.9 & 112.4\\
         PIC \cite{fang2024explore} & ~ & 69.2 & 64.7 & 58.4 & 72.5 & 77.4 & 62.8 & 72.0 & 65.4 & 60.3 & 71.8 & 79.5 & 60.3 & 70.2 & 60.8 & 54.9 & 71.8 & 78.3 & 60.6\\
    \hline
         AdaBN \cite{li2016revisiting} & \multirow{7}{*}{\makecell{CoTTA\\Models}}& 58.7 & 52.1 & 37.7 & 64.1 & 76.8 & 57.2 & 58.9 & 51.5 & 37.2 & 64.1 & 74.2 & 53.9 & 56.8 & 50.3 & 35.5 & 62.1 & 71.7 & 51.1\\
         TENT \cite{wang2021tent} & ~ & 57.9 & 50.6 & 36.8 & 64.8 & 76.4 & 55.0 & 57.8 & 50.0 & 36.7 & 64.7 & 73.5 & 51.1 & 55.2 & 48.4 & 35.0 & 62.1 & 69.2 & 49.7\\
         CoTTA \cite{wang2022continual} & ~ & 58.3 & 50.1 & 36.4 & 62.5 & 73.6 & 50.3 & 56.7 & 49.0 & 34.4 & 60.4 & 71.8 & 49.1 & 55.2 & 46.9 & 34.3 & 59.6 & 66.3 & 48.5\\
         ViDA \cite{liu2023vida} & ~ & 52.4 & 47.2 & 35.1 & 58.2 & 69.8 & 47.5 & 51.6 & 46.9 & 34.3 & 57.6 & 67.2 & 45.5 & 51.3 & 46.2 & 32.8 & 54.4 & 63.1 & 42.8\\
         RMT \cite{dobler2023robust} & ~ &31.2 & 44.0 & 34.3 & 47.4 & 59.6 & 39.9 & 30.6 & 43.5 & 33.9 & 45.6 & 53.0 & 35.8 & 30.4 & 42.7 & 33.8 & 45.9 & 51.1 & 36.4\\
         SANTA \cite{chakrabarty2023santa} & ~ & 32.3 & 42.1 & 37.8 & 44.9 & 55.2 & 38.6 & 31.7 & 41.9 & 37.4 & 42.0 & 53.4 & 35.6 & 30.1 & 41.6 & 36.4 & 40.6 & 52.9 & 34.7\\
         Our \textbf{PCoTTA} & ~ & \textbf{6.3} & \textbf{21.4} & \textbf{15.4} & \textbf{8.9} & \textbf{28.3} & \textbf{20.7} & \textbf{5.5} & \textbf{19.9} &\textbf{14.6} & \textbf{8.5} & \textbf{26.9} & \textbf{19.6} & \textbf{5.4} & \textbf{18.6} & \textbf{12.1} & \textbf{8.2} & \textbf{25.2} & \textbf{19.3}\\
    \hline
    \end{tabular}}
    \label{tab:compared_method}
\end{table}

\subsection{Main Results}
Table \ref{tab:compared_method} shows the comparison results of our PCoTTA against other methods across tasks of reconstruction, denoising, and registration in the introduced setting. Our method consistently outperforms others by a large margin, demonstrating superior adaptability in a multi-domain multi-task setting. 
Conventional methods such as PointNet~\cite{qi2017pointnet}, DGCNN~\cite{wang2019dynamic}, and PCT~\cite{guo2021pct} often struggle with unseen data, leading to significant performance drops. Augmentation-based methods like Pointmixup~\cite{chen2020pointmixup} and PointCutMix~\cite{zhang2022pointcutmix}, though adapted for multi-domain learning, exhibit limited performance in multi-task generalization. Despite incorporating task-specific heads, these methods still fall short compared to our unified model, which excels across all tasks due to our Automatic Prototype Mixture (APM) and Gaussian Splatted Feature Shifting (GSFS) modules.
While PIC \cite{fang2024explore} performs well in multi-task scenarios, its transferability is limited, often failing with changing target data. Our PCoTTA addresses this by aligning the target data with source prototypes and dynamically updating learnable prototypes at test time, effectively narrowing the domain gap. Our method demonstrates strong continuous online learning abilities, improving results across multiple validations and showing resilience against catastrophic forgetting and error accumulation.

We compare our PCoTTA with advanced CoTTA methods like AdaBN~\cite{li2016revisiting}, TENT~\cite{wang2021tent}, CoTTA~\cite{wang2022continual}, ViDA~\cite{liu2023vida}, RMT~\cite{dobler2023robust}, and SANTA~\cite{chakrabarty2023santa}. 
To ensure fairness, we also update the LayerNorm parameters for AdaBN, TENT, and SANTA. While these methods handle continuously changing targets, they struggle with multi-task aspect in our challenging setting. Even when equipped with multi-task capabilities, these methods still underperform compared to our PCoTTA. Our success is attributed to three main factors: (1) Usually, these methods heavily rely on the student-teacher architecture to realize consistency regularization. As a result, they would inevitably introduce pseudo-label noise, leading to error accumulation. Although they use symmetric cross-entropy or other techniques to alleviate the pseudo-label noise, such problems still exist and cannot be fundamentally addressed. In contrast, our PCoTTA framework does not use any online or offline pseudo-labeling techniques, which inherently avoids the risk of error accumulation. (2) These methods are specifically designed for CoTTA in 2D images and perform well on 2D images. However, compared to 2D images, 3D point cloud data is disordered, unstructured, and sparsely distributed, making these 2D image-based CoTTA methods less effective or even inapplicable. Our method involves specific designs for 3D point cloud data, \textit{e.g.,} Gaussian Splatted-based Graph Attention for comprehensive, patch similarity-based adaptation, well-suited for 3D data, and achieves better performances than these methods. (3) These methods often focus on single tasks and all lack specialized design in multi-task learning, which may lead to gradient conflicts in the optimization process of continual test-time adaptation. Instead, our PCoTTA devises task-specific prototype banks where individual source-learnable prototype pairs are used for different adaptations in each task, thus favoring the multi-task learning in our setting.

\subsection{Ablation Studies}

\noindent\textbf{Effect of Each Component.} 
Table \ref{tab:ablation} shows the effects of different components.
Compared with the baseline, Model A  simply shifts target features by equally fusing with every source-learnable prototype pair (APM), demonstrating that our prototype bank effectively enriches the source information for targets, thereby improving the model's transferability. By adding GSFS, we achieve better performance. 
This is because Model B uses the similarity between targets and prototypes as weights during aggregation, and meanwhile, our attention mechanism in GSFS also enables dynamic updating of these weights, offering greater weights to prototypes closer to the current sample. 
Finally, adding CPR (Ours) enables the prototype bank's learnable prototypes to be more distinct, achieving the best performance. These improvements confirm that these individual components are complementary
and together they significantly promote the performance.

\noindent\textbf{Quantity of Learnable Prototypes.} 
We conducted an additional ablation study on the number of learnable prototypes, as shown in Table \ref{tab:num_prototype}, and the results indicate marginal changes. Additionally, we show the case with no learnable prototypes (\textit{i.e.,} quantity 0), where our method degrades to aligning the target feature by solely considering source prototypes’ similarities. While this case achieves some degree of test-time adaptation, its performance is less decent than our PCoTTA.

\begin{table}[htbp]
	\begin{minipage}[c]{0.52\columnwidth}
        \caption{Ablation studies on our proposed three modules. We report the CD ($\times 10^{-3}$) for three different tasks on ModelNet40. }
        \centering
        \begin{tabular}{l|ccc|ccc}
        \hline
         Models & APM & GSFS & CPR & Rec. & Den. & Reg.\\
        \hline
         Baseline & & & & 78.9 & 123.6 & 110.6\\
         A & \checkmark & & & 23.5 & 37.4 & 30.1\\
         B & \checkmark & \checkmark & & 14.6 & 31.2 & 27.2\\
         \textbf{Ours} & \checkmark & \checkmark & \checkmark & \textbf{5.4} & \textbf{18.6} & \textbf{12.1}\\
        \hline
        \end{tabular}
		\label{tab:ablation}
	\end{minipage}
    \hspace{3pt}
	\begin{minipage}[c]{0.43\columnwidth}
		\centering
            \caption{Ablation studies on the quantity of learnable prototypes.}
            \centering
            \begin{tabular}{l|c|ccc}
            \hline
            Models & Quantity & Rec. & Den. & Reg.\\
            \hline
            I & 0 & 15.5 & 32.4 & 30.7\\
            II & 1 & 8.2 & 24.9 & 17.4\\
            \textbf{Ours} & \textbf{2} & \textbf{5.4} & \textbf{18.6} & \textbf{12.1}\\
            III & 3 & 6.8 & 19.5 & 14.8\\
            IV & 4 & 6.5 & 20.3 & 14.0\\
            \hline
            \end{tabular}
		\label{tab:num_prototype}
	\end{minipage}
 \end{table}

\noindent\textbf{Cross Validation.}
Table \ref{tab:cross} shows our model's effectiveness in bridging the domain gap from the synthetic to real scan data. 
Consistently, our method surpasses CoTTA~\cite{wang2022continual} in all tasks, demonstrating the superiority of our method. Remarkably, our model also performs better than CoTTA \cite{wang2022continual} when pre-trained on the two real scan datasets which involve background interference and missing parts. This underscores our method's strong transferability between various domains. 

\noindent\textbf{Efficiency Analysis.}
We present an analysis of model parameters and running time in Table \ref{tab:runtime}. The results show that our method achieves fast inference on target data, and our model has the fewest parameters compared to other CTTA methods. As such, this shows potential for many real-world applications, \textit{e.g.,} autonomous driving and virtual reality, since our PCoTTA is an end-to-end test-time adaptation method without relying on a teacher-student model or pseudo-labeling technique, it is more efficient and suitable for real-time deployment.

\begin{table}[htbp]
	\begin{minipage}[c]{0.52\columnwidth}
        \caption{Cross validation with synthetic data: ShapeNet (SP), ModelNet40 (MN), and real scan data: ScanNet (SN), ScanObjectNN (SO). }
        \centering
        \begin{tabular}{l|c|ccc}
        \hline
        Methods & Sources $\rightarrow$ Targets & Rec. & Den. & Reg.\\
        \hline
        CoTTA \cite{wang2022continual} & \multirow{2}{*}{SP + MN $\rightarrow$ SN + SO} & 63.6 & 70.4 & 57.2\\
        \textbf{Ours} & ~ & \textbf{12.7} & \textbf{30.7} & \textbf{23.2} \\
        \hline
        CoTTA \cite{wang2022continual} & \multirow{2}{*}{SN + SO $\rightarrow$ SP + MN} & 58.8 & 50.3 & 39.6\\
        \textbf{Ours} & ~ & \textbf{10.4} & \textbf{26.1} & \textbf{17.4}\\
        \hline
        \end{tabular}
        \label{tab:cross}
	\end{minipage}
    \hspace{3pt}
	\begin{minipage}[c]{0.43\columnwidth}
        \caption{Comparison of model efficiency. We report the Runtime (s), Flops (G), and Parameters (M) as metrics.}
        \centering
        \begin{tabular}{lccc}
        \hline
         Methods & Run. & Flop. & Para.\\
        \hline
        CoTTA \cite{wang2022continual} & 4.96 & 24.26 & 86.72\\
        ViDA \cite{liu2023vida}& 5.14 & 19.99 & 100.98\\
        \textbf{Ours} & \textbf{0.06} & \textbf{12.11} & \textbf{28.91} \\
        \hline
        \end{tabular}
        \label{tab:runtime}
	\end{minipage}
 \end{table}

\subsection{Visualization and Analysis}

\begin{figure}[t]
    \centering
    \includegraphics[width=1\linewidth]{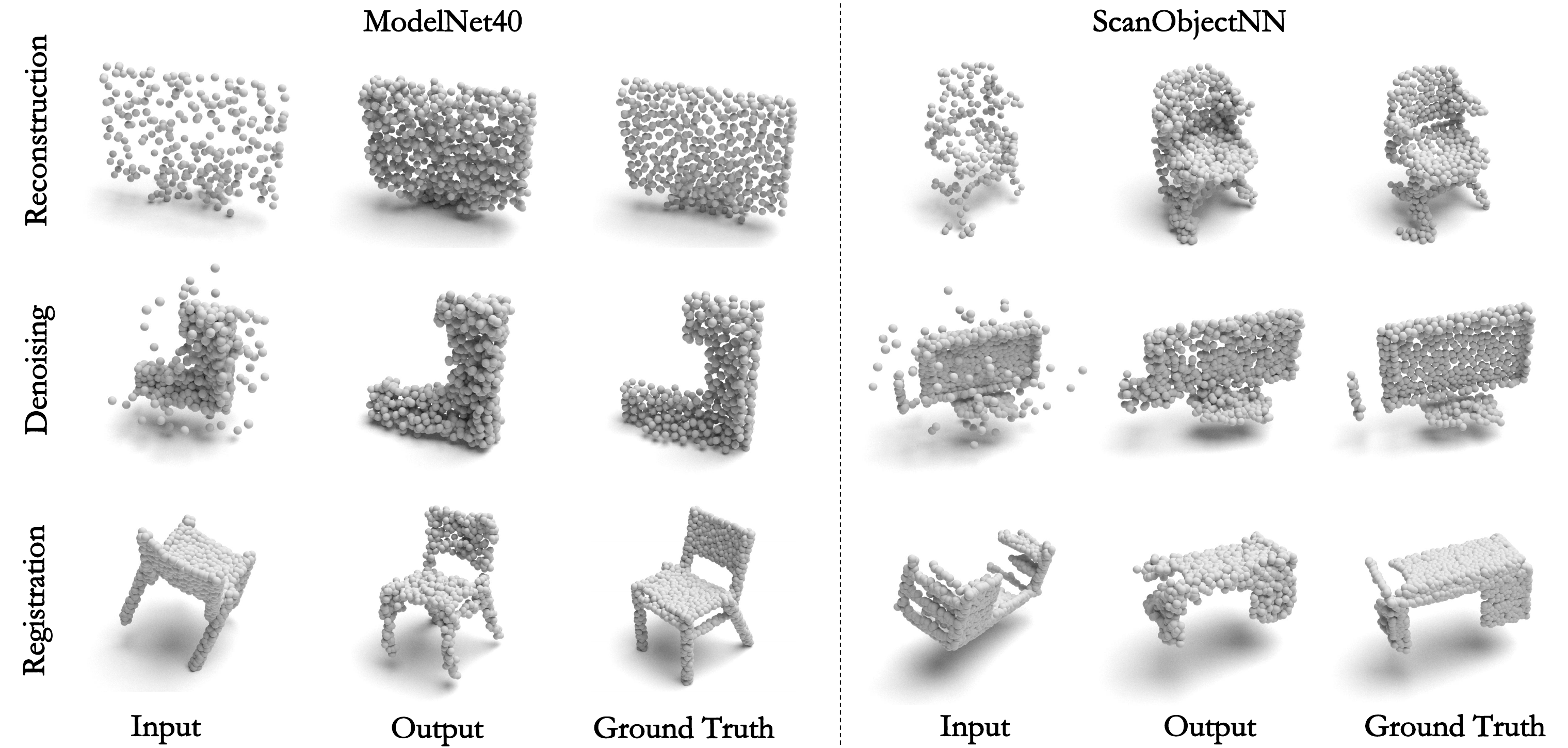}
    \caption{Visualization of our PCoTTA's prediction and their ground truths under $3$ different tasks. }
    \label{fig:visual_results}
\end{figure}

\noindent\textbf{Visualization of Different Tasks.} Figure \ref{fig:visual_results} illustrates the qualitative results in the last round of our PCoTTA model. 
From the figure, we have two observations. Firstly, our proposed PCoTTA manages to generate quality predictions in the continually changing target domain by leveraging the proposed distinctive prototype bank, minimizing the discrepancies between source and target domains. Secondly, without retraining a CoTTA method for each task, our proposed PCoTTA is able to successfully handle multiple tasks such as point cloud reconstruction, denoising, and registration and multiple domains with a unified model, demonstrating strong practicability and transferability in the real world. We provide more visual comparisons with state-of-the-art methods in Appendix \ref{supp:vis_results}.

\begin{figure}[htb]
    \centering
    \includegraphics[width=1\linewidth]{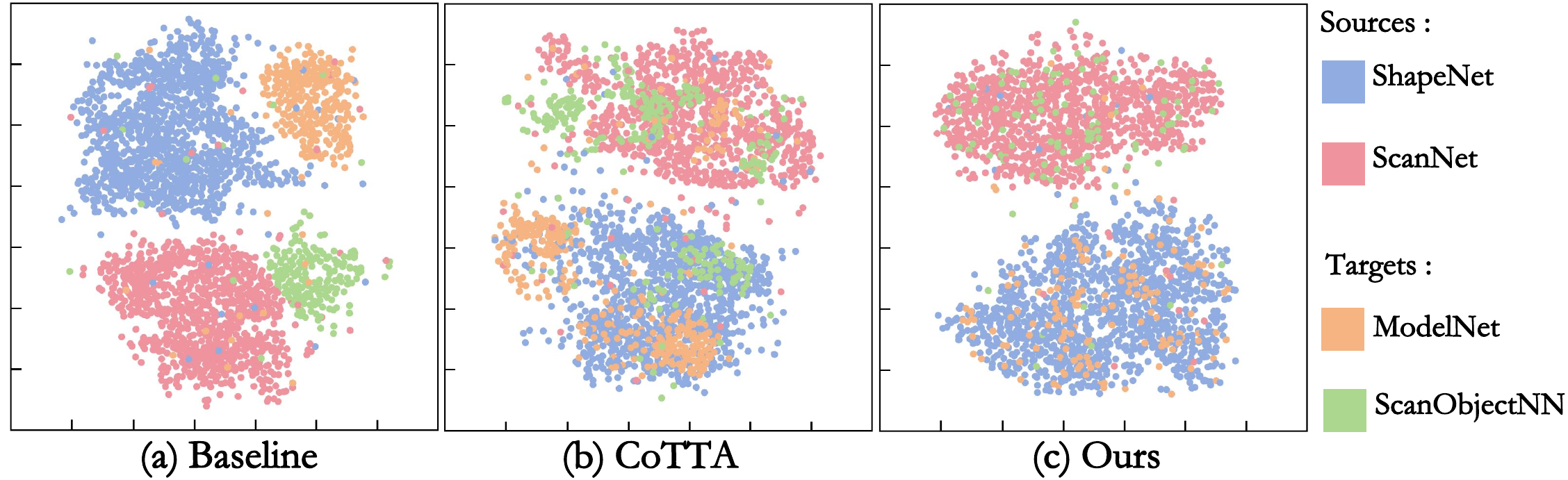}
    \caption{T-SNE visualization of the source and target features. }
    \label{fig:tsne}
\end{figure}

\noindent\textbf{T-SNE Feature Visualization.
} To understand how our PCoTTA aligns the domains, we visualize the feature distributions of the source and target domains via t-SNE. We display the latent features of the point cloud reconstruction task in Figure \ref{fig:tsne}. 
From the figure, we make the following observations: The baseline model means directly deploying the source pre-trained model in the continually changing domains, resulting in an unsatisfactory alignment. Although CoTTA~\cite{wang2022continual} aligns the source and target domains to some extent, there still exists some cases of miss-alignment or over-alignment. For example, some samples are either not aligned with the cluster or over-clustered. In contrast, our PCoTTA achieves a better and more even feature alignment across domains, demonstrating its superiority in narrowing domain shifts in continually changing environments.

\section{Conclusion}
\label{sec:conclusion}
In this paper, we present an innovative, pioneering, and unified framework, namely PCoTTA for Continual Test-Time Adaptation in multi-task point cloud understanding, 
boosting the model's transferability towards the continually changing target domains. 
Our approach effectively mitigates catastrophic forgetting and error accumulation issues through the three novel modules: automatic prototype mixture (APM), Gaussian Splatted feature shifting (GSFS), and contrastive prototype repulsion (CPR). These three components make our model more adaptable and robust across continually changing domains by aligning the targets towards all source domains. 
Furthermore, we present a new benchmark in terms of the practical Continual Test-Time Adaptation for multi-task point cloud understanding.
Comprehensive experiments show our PCoTTA's superior performance, proving its efficacy in significantly improving the model's transferability across various domains. We believe our work will inspire a new direction and interesting ideas in the community, in terms of Continual Test-Time Adaptation for multi-task point cloud understanding. 

\clearpage
\section*{Acknowledgments}
Jincen Jiang is supported by the China Scholarship Council (Grand Number 202306300023), and the Research and Development Fund of Bournemouth University. 

{\small
  \bibliographystyle{ieee_fullname}
  \bibliography{refbib}

\begin{thebibliography}{10}\itemsep=-1pt

\bibitem{achituve2021self}
Idan Achituve, Haggai Maron, and Gal Chechik.
\newblock Self-supervised learning for domain adaptation on point clouds.
\newblock In {\em Proceedings of the IEEE/CVF Winter Conference on Applications of Computer Vision}, pages 123--133, 2021.

\bibitem{brahma2023probabilistic}
Dhanajit Brahma and Piyush Rai.
\newblock A probabilistic framework for lifelong test-time adaptation.
\newblock In {\em Proceedings of the IEEE/CVF Conference on Computer Vision and Pattern Recognition}, pages 3582--3591, 2023.

\bibitem{cao2023multi}
Haozhi Cao, Yuecong Xu, Jianfei Yang, Pengyu Yin, Shenghai Yuan, and Lihua Xie.
\newblock Multi-modal continual test-time adaptation for 3d semantic segmentation.
\newblock In {\em Proceedings of the IEEE/CVF International Conference on Computer Vision}, pages 18809--18819, 2023.

\bibitem{chakrabarty2023santa}
Goirik Chakrabarty, Manogna Sreenivas, and Soma Biswas.
\newblock Santa: Source anchoring network and target alignment for continual test time adaptation.
\newblock {\em Transactions on Machine Learning Research}, 2023.

\bibitem{chang2015shapenet}
Angel~X Chang, Thomas Funkhouser, Leonidas Guibas, Pat Hanrahan, Qixing Huang, Zimo Li, Silvio Savarese, Manolis Savva, Shuran Song, Hao Su, et~al.
\newblock Shapenet: An information-rich 3d model repository.
\newblock {\em arXiv preprint arXiv:1512.03012}, 2015.

\bibitem{chen2020pointmixup}
Yunlu Chen, Vincent~Tao Hu, Efstratios Gavves, Thomas Mensink, Pascal Mettes, Pengwan Yang, and Cees~GM Snoek.
\newblock Pointmixup: Augmentation for point clouds.
\newblock In {\em Computer Vision--ECCV 2020: 16th European Conference, Glasgow, UK, August 23--28, 2020, Proceedings, Part III 16}, pages 330--345. Springer, 2020.

\bibitem{dai2017scannet}
Angela Dai, Angel~X Chang, Manolis Savva, Maciej Halber, Thomas Funkhouser, and Matthias Nie{\ss}ner.
\newblock Scannet: Richly-annotated 3d reconstructions of indoor scenes.
\newblock In {\em Proceedings of the IEEE/CVF Conference on Computer Vision and Pattern Recognition}, pages 5828--5839, 2017.

\bibitem{de2024straightpcf}
Dasith de Silva~Edirimuni, Xuequan Lu, Gang Li, Lei Wei, Antonio Robles-Kelly, and Hongdong Li.
\newblock Straightpcf: Straight point cloud filtering.
\newblock In {\em Proceedings of the IEEE/CVF Conference on Computer Vision and Pattern Recognition}, pages 20721--20730, 2024.

\bibitem{de2023iterativepfn}
Dasith de Silva~Edirimuni, Xuequan Lu, Zhiwen Shao, Gang Li, Antonio Robles-Kelly, and Ying He.
\newblock Iterativepfn: True iterative point cloud filtering.
\newblock In {\em Proceedings of the IEEE/CVF Conference on Computer Vision and Pattern Recognition}, pages 13530--13539, 2023.

\bibitem{dobler2023robust}
Mario D{\"o}bler, Robert~A Marsden, and Bin Yang.
\newblock Robust mean teacher for continual and gradual test-time adaptation.
\newblock In {\em Proceedings of the IEEE/CVF Conference on Computer Vision and Pattern Recognition}, pages 7704--7714, 2023.

\bibitem{fang2024explore}
Zhongbin Fang, Xiangtai Li, Xia Li, Joachim~M Buhmann, Chen~Change Loy, and Mengyuan Liu.
\newblock Explore in-context learning for 3d point cloud understanding.
\newblock {\em Advances in Neural Information Processing Systems}, 36, 2024.

\bibitem{fung2024semreg}
Sheldon Fung, Xuequan Lu, D Edirimuni, Wei Pan, Xiao Liu, and Hongdong Li.
\newblock Semreg: Semantics constrained point cloud registration.
\newblock In {\em Proceedings of the Europeon Conference on Computer Vision}, 2024.

\bibitem{gan2023decorate}
Yulu Gan, Yan Bai, Yihang Lou, Xianzheng Ma, Renrui Zhang, Nian Shi, and Lin Luo.
\newblock Decorate the newcomers: Visual domain prompt for continual test time adaptation.
\newblock In {\em Proceedings of the AAAI Conference on Artificial Intelligence}, volume~37, pages 7595--7603, 2023.

\bibitem{gong2022note}
Taesik Gong, Jongheon Jeong, Taewon Kim, Yewon Kim, Jinwoo Shin, and Sung-Ju Lee.
\newblock Note: Robust continual test-time adaptation against temporal correlation.
\newblock {\em Advances in Neural Information Processing Systems}, 35:27253--27266, 2022.

\bibitem{guo2021pct}
Meng-Hao Guo, Jun-Xiong Cai, Zheng-Ning Liu, Tai-Jiang Mu, Ralph~R Martin, and Shi-Min Hu.
\newblock Pct: Point cloud transformer.
\newblock {\em Computational Visual Media}, 7:187--199, 2021.

\bibitem{hatem2023point}
Ahmed Hatem, Yiming Qian, and Yang Wang.
\newblock Point-tta: Test-time adaptation for point cloud registration using multitask meta-auxiliary learning.
\newblock In {\em Proceedings of the IEEE/CVF International Conference on Computer Vision}, pages 16494--16504, 2023.

\bibitem{hatem2023test}
Ahmed Hatem, Yiming Qian, and Yang Wang.
\newblock Test-time adaptation for point cloud upsampling using meta-learning.
\newblock In {\em 2023 IEEE/RSJ International Conference on Intelligent Robots and Systems (IROS)}, pages 1284--1291. IEEE, 2023.

\bibitem{hu2023density}
Qianjiang Hu, Daizong Liu, and Wei Hu.
\newblock Density-insensitive unsupervised domain adaption on 3d object detection.
\newblock In {\em Proceedings of the IEEE/CVF Conference on Computer Vision and Pattern Recognition}, pages 17556--17566, 2023.

\bibitem{jiang2023masked}
Jincen Jiang, Xuequan Lu, Lizhi Zhao, Richard Dazaley, and Meili Wang.
\newblock Masked autoencoders in 3d point cloud representation learning.
\newblock {\em IEEE Transactions on Multimedia}, 2023.

\bibitem{jiang2024dhgcn}
Jincen Jiang, Lizhi Zhao, Xuequan Lu, Wei Hu, Imran Razzak, and Meili Wang.
\newblock Dhgcn: Dynamic hop graph convolution network for self-supervised point cloud learning.
\newblock In {\em Proceedings of the AAAI Conference on Artificial Intelligence}, volume~38, pages 12883--12891, 2024.

\bibitem{jiang2025dg}
Jincen Jiang, Qianyu Zhou, Yuhang Li, Xuequan Lu, Meili Wang, Lizhuang Ma, Jian Chang, and Jian~Jun Zhang.
\newblock Dg-pic: Domain generalized point-in-context learning for point cloud understanding.
\newblock In {\em European Conference on Computer Vision}, pages 455--474. Springer, 2025.

\bibitem{jiang2021lidarnet}
Peng Jiang and Srikanth Saripalli.
\newblock Lidarnet: A boundary-aware domain adaptation model for point cloud semantic segmentation.
\newblock In {\em 2021 IEEE International Conference on Robotics and Automation (ICRA)}, pages 2457--2464. IEEE, 2021.

\bibitem{li2016revisiting}
Yanghao Li, Naiyan Wang, Jianping Shi, Jiaying Liu, and Xiaodi Hou.
\newblock Revisiting batch normalization for practical domain adaptation.
\newblock {\em arXiv preprint arXiv:1603.04779}, 2016.

\bibitem{liang2022point}
Hanxue Liang, Hehe Fan, Zhiwen Fan, Yi Wang, Tianlong Chen, Yu Cheng, and Zhangyang Wang.
\newblock Point cloud domain adaptation via masked local 3d structure prediction.
\newblock In {\em European Conference on Computer Vision}, pages 156--172. Springer, 2022.

\bibitem{liu2023vida}
Jiaming Liu, Senqiao Yang, Peidong Jia, Renrui Zhang, Ming Lu, Yandong Guo, Wei Xue, and Shanghang Zhang.
\newblock Vida: Homeostatic visual domain adapter for continual test time adaptation.
\newblock {\em arXiv preprint arXiv:2306.04344}, 2023.

\bibitem{liu2021adversarial}
Wei Liu, Zhiming Luo, Yuanzheng Cai, Ying Yu, Yang Ke, Jos{\'e}~Marcato Junior, Wesley~Nunes Gon{\c{c}}alves, and Jonathan Li.
\newblock Adversarial unsupervised domain adaptation for 3d semantic segmentation with multi-modal learning.
\newblock {\em ISPRS Journal of Photogrammetry and Remote Sensing}, 176:211--221, 2021.

\bibitem{long2024dgmamba}
Shaocong Long, Qianyu Zhou, Xiangtai Li, Xuequan Lu, Chenhao Ying, Yuan Luo, Lizhuang Ma, and Shuicheng Yan.
\newblock Dgmamba: Domain generalization via generalized state space model.
\newblock In {\em Proceedings of the 30th ACM International Conference on Multimedia (ACM MM)}, 2024.

\bibitem{loshchilov2019decoupled}
Ilya Loshchilov and Frank Hutter.
\newblock Decoupled weight decay regularization.
\newblock In {\em International Conference on Learning Representations}, 2019.

\bibitem{luo2021diffusion}
Shitong Luo and Wei Hu.
\newblock Diffusion probabilistic models for 3d point cloud generation.
\newblock In {\em Proceedings of the IEEE/CVF Conference on Computer Vision and Pattern Recognition}, pages 2837--2845, 2021.

\bibitem{luo2021unsupervised}
Zhipeng Luo, Zhongang Cai, Changqing Zhou, Gongjie Zhang, Haiyu Zhao, Shuai Yi, Shijian Lu, Hongsheng Li, Shanghang Zhang, and Ziwei Liu.
\newblock Unsupervised domain adaptive 3d detection with multi-level consistency.
\newblock In {\em Proceedings of the IEEE/CVF International Conference on Computer Vision}, pages 8866--8875, 2021.

\bibitem{niloy2024effective}
Fahim~Faisal Niloy, Sk~Miraj Ahmed, Dripta~S Raychaudhuri, Samet Oymak, and Amit~K Roy-Chowdhury.
\newblock Effective restoration of source knowledge in continual test time adaptation.
\newblock In {\em Proceedings of the IEEE/CVF Winter Conference on Applications of Computer Vision}, pages 2091--2100, 2024.

\bibitem{niu2022efficient}
Shuaicheng Niu, Jiaxiang Wu, Yifan Zhang, Yaofo Chen, Shijian Zheng, Peilin Zhao, and Mingkui Tan.
\newblock Efficient test-time model adaptation without forgetting.
\newblock In {\em International Conference on Machine Learning}, pages 16888--16905. PMLR, 2022.

\bibitem{pang2022masked}
Yatian Pang, Wenxiao Wang, Francis~EH Tay, Wei Liu, Yonghong Tian, and Li Yuan.
\newblock Masked autoencoders for point cloud self-supervised learning.
\newblock In {\em European Conference on Computer Vision}, pages 604--621. Springer, 2022.

\bibitem{qi2017pointnet}
Charles~R Qi, Hao Su, Kaichun Mo, and Leonidas~J Guibas.
\newblock Pointnet: Deep learning on point sets for 3d classification and segmentation.
\newblock In {\em Proceedings of the IEEE/CVF Conference on Computer Vision and Pattern Recognition}, pages 652--660, 2017.

\bibitem{qi2017pointnet++}
Charles~Ruizhongtai Qi, Li Yi, Hao Su, and Leonidas~J Guibas.
\newblock Pointnet++: Deep hierarchical feature learning on point sets in a metric space.
\newblock {\em Advances in Neural Information Processing Systems}, 30, 2017.

\bibitem{qin2019pointdan}
Can Qin, Haoxuan You, Lichen Wang, C-C~Jay Kuo, and Yun Fu.
\newblock Pointdan: A multi-scale 3d domain adaption network for point cloud representation.
\newblock {\em Advances in Neural Information Processing Systems}, 32, 2019.

\bibitem{shaban2023lidar}
Amirreza Shaban, JoonHo Lee, Sanghun Jung, Xiangyun Meng, and Byron Boots.
\newblock Lidar-uda: Self-ensembling through time for unsupervised lidar domain adaptation.
\newblock In {\em Proceedings of the IEEE/CVF International Conference on Computer Vision}, pages 19784--19794, 2023.

\bibitem{shao2024trici}
Di Shao, Xuequan Lu, Weijia Wang, Xiao Liu, and Ajmal~Saeed Mian.
\newblock Trici: Triple cross-intra branch contrastive learning for point cloud analysis.
\newblock {\em IEEE Transactions on Visualization and Computer Graphics}, 2024.

\bibitem{shen2022domain}
Yuefan Shen, Yanchao Yang, Mi Yan, He Wang, Youyi Zheng, and Leonidas~J Guibas.
\newblock Domain adaptation on point clouds via geometry-aware implicits.
\newblock In {\em Proceedings of the IEEE/CVF Conference on Computer Vision and Pattern Recognition}, pages 7223--7232, 2022.

\bibitem{sojka2023ar}
Damian S{\'o}jka, Sebastian Cygert, Bart{\l}omiej Twardowski, and Tomasz Trzci{\'n}ski.
\newblock Ar-tta: A simple method for real-world continual test-time adaptation.
\newblock In {\em Proceedings of the IEEE/CVF International Conference on Computer Vision}, pages 3491--3495, 2023.

\bibitem{thomas2019kpconv}
Hugues Thomas, Charles~R Qi, Jean-Emmanuel Deschaud, Beatriz Marcotegui, Fran{\c{c}}ois Goulette, and Leonidas~J Guibas.
\newblock Kpconv: Flexible and deformable convolution for point clouds.
\newblock In {\em Proceedings of the IEEE/CVF International Conference on Computer Vision}, pages 6411--6420, 2019.

\bibitem{uy2019revisiting}
Mikaela~Angelina Uy, Quang-Hieu Pham, Binh-Son Hua, Thanh Nguyen, and Sai-Kit Yeung.
\newblock Revisiting point cloud classification: A new benchmark dataset and classification model on real-world data.
\newblock In {\em Proceedings of the IEEE/CVF International Conference on Computer Vision}, pages 1588--1597, 2019.

\bibitem{wang2021tent}
Dequan Wang, Evan Shelhamer, Shaoteng Liu, Bruno Olshausen, and Trevor Darrell.
\newblock Tent: Fully test-time adaptation by entropy minimization.
\newblock In {\em International Conference on Learning Representations}, 2021.

\bibitem{wang2021cross}
Feiyu Wang, Wen Li, and Dong Xu.
\newblock Cross-dataset point cloud recognition using deep-shallow domain adaptation network.
\newblock {\em IEEE Transactions on Image Processing}, 30:7364--7377, 2021.

\bibitem{wang2022continual}
Qin Wang, Olga Fink, Luc Van~Gool, and Dengxin Dai.
\newblock Continual test-time domain adaptation.
\newblock In {\em Proceedings of the IEEE/CVF Conference on Computer Vision and Pattern Recognition}, pages 7201--7211, 2022.

\bibitem{wang2024continual}
Yanshuo Wang, Jie Hong, Ali Cheraghian, Shafin Rahman, David Ahmedt-Aristizabal, Lars Petersson, and Mehrtash Harandi.
\newblock Continual test-time domain adaptation via dynamic sample selection.
\newblock In {\em Proceedings of the IEEE/CVF Winter Conference on Applications of Computer Vision}, pages 1701--1710, 2024.

\bibitem{wang2019dynamic}
Yue Wang, Yongbin Sun, Ziwei Liu, Sanjay~E Sarma, Michael~M Bronstein, and Justin~M Solomon.
\newblock Dynamic graph cnn for learning on point clouds.
\newblock {\em ACM Transactions on Graphics}, 38(5):1--12, 2019.

\bibitem{wang2023ssda3d}
Yan Wang, Junbo Yin, Wei Li, Pascal Frossard, Ruigang Yang, and Jianbing Shen.
\newblock Ssda3d: Semi-supervised domain adaptation for 3d object detection from point cloud.
\newblock In {\em Proceedings of the AAAI Conference on Artificial Intelligence}, volume~37, pages 2707--2715, 2023.

\bibitem{wei2022lidar}
Yi Wei, Zibu Wei, Yongming Rao, Jiaxin Li, Jie Zhou, and Jiwen Lu.
\newblock Lidar distillation: Bridging the beam-induced domain gap for 3d object detection.
\newblock In {\em European Conference on Computer Vision}, pages 179--195. Springer, 2022.

\bibitem{wu2023scoda}
Yushuang Wu, Zizheng Yan, Ce Chen, Lai Wei, Xiao Li, Guanbin Li, Yihao Li, Shuguang Cui, and Xiaoguang Han.
\newblock Scoda: Domain adaptive shape completion for real scans.
\newblock In {\em Proceedings of the IEEE/CVF Conference on Computer Vision and Pattern Recognition}, pages 17630--17641, 2023.

\bibitem{wu20153d}
Zhirong Wu, Shuran Song, Aditya Khosla, Fisher Yu, Linguang Zhang, Xiaoou Tang, and Jianxiong Xiao.
\newblock 3d shapenets: A deep representation for volumetric shapes.
\newblock In {\em Proceedings of the IEEE/CVF Conference on Computer Vision and Pattern Recognition}, pages 1912--1920, 2015.

\bibitem{xu2021paconv}
Mutian Xu, Runyu Ding, Hengshuang Zhao, and Xiaojuan Qi.
\newblock Paconv: Position adaptive convolution with dynamic kernel assembling on point clouds.
\newblock In {\em Proceedings of the IEEE/CVF Conference on Computer Vision and Pattern Recognition}, pages 3173--3182, 2021.

\bibitem{yang2021st3d}
Jihan Yang, Shaoshuai Shi, Zhe Wang, Hongsheng Li, and Xiaojuan Qi.
\newblock St3d: Self-training for unsupervised domain adaptation on 3d object detection.
\newblock In {\em Proceedings of the IEEE/CVF Conference on Computer Vision and Pattern Recognition}, pages 10368--10378, 2021.

\bibitem{yang2022st3d++}
Jihan Yang, Shaoshuai Shi, Zhe Wang, Hongsheng Li, and Xiaojuan Qi.
\newblock St3d++: Denoised self-training for unsupervised domain adaptation on 3d object detection.
\newblock {\em IEEE Transactions on Pattern Analysis and Machine Intelligence}, 45(5):6354--6371, 2022.

\bibitem{yang2024exploring}
Senqiao Yang, Jiarui Wu, Jiaming Liu, Xiaoqi Li, Qizhe Zhang, Mingjie Pan, Yulu Gan, Zehui Chen, and Shanghang Zhang.
\newblock Exploring sparse visual prompt for domain adaptive dense prediction.
\newblock In {\em Proceedings of the AAAI Conference on Artificial Intelligence}, volume~38, pages 16334--16342, 2024.

\bibitem{yi2021complete}
Li Yi, Boqing Gong, and Thomas Funkhouser.
\newblock Complete \& label: A domain adaptation approach to semantic segmentation of lidar point clouds.
\newblock In {\em Proceedings of the IEEE/CVF Conference on Computer Vision and Pattern Recognition}, pages 15363--15373, 2021.

\bibitem{yu2022point}
Xumin Yu, Lulu Tang, Yongming Rao, Tiejun Huang, Jie Zhou, and Jiwen Lu.
\newblock Point-bert: Pre-training 3d point cloud transformers with masked point modeling.
\newblock In {\em Proceedings of the IEEE/CVF Conference on Computer Vision and Pattern Recognition}, pages 19313--19322, 2022.

\bibitem{yu2023noise}
Zhiqi Yu, Jingjing Li, Zhekai Du, Fengling Li, Lei Zhu, and Yang Yang.
\newblock Noise-robust continual test-time domain adaptation.
\newblock In {\em Proceedings of the 31st ACM International Conference on Multimedia}, pages 2654--2662, 2023.

\bibitem{zhang2022pointcutmix}
Jinlai Zhang, Lyujie Chen, Bo Ouyang, Binbin Liu, Jihong Zhu, Yujin Chen, Yanmei Meng, and Danfeng Wu.
\newblock Pointcutmix: Regularization strategy for point cloud classification.
\newblock {\em Neurocomputing}, 505:58--67, 2022.

\bibitem{zhang2021srdan}
Weichen Zhang, Wen Li, and Dong Xu.
\newblock Srdan: Scale-aware and range-aware domain adaptation network for cross-dataset 3d object detection.
\newblock In {\em Proceedings of the IEEE/CVF Conference on Computer Vision and Pattern Recognition}, pages 6769--6779, 2021.

\bibitem{zhao2021epointda}
Sicheng Zhao, Yezhen Wang, Bo Li, Bichen Wu, Yang Gao, Pengfei Xu, Trevor Darrell, and Kurt Keutzer.
\newblock epointda: An end-to-end simulation-to-real domain adaptation framework for lidar point cloud segmentation.
\newblock In {\em Proceedings of the AAAI Conference on Artificial Intelligence}, volume~35, pages 3500--3509, 2021.

\bibitem{zhou2023context}
Qianyu Zhou, Zhengyang Feng, Qiqi Gu, Jiangmiao Pang, Guangliang Cheng, Xuequan Lu, Jianping Shi, and Lizhuang Ma.
\newblock Context-aware mixup for domain adaptive semantic segmentation.
\newblock {\em IEEE Transactions on Circuits and Systems for Video Technology (TCSVT)}, 33(2):804--817, 2023.

\bibitem{zhou2023self}
Qianyu Zhou, Qiqi Gu, Jiangmiao Pang, Xuequan Lu, and Lizhuang Ma.
\newblock Self-adversarial disentangling for specific domain adaptation.
\newblock {\em IEEE Transactions on Pattern Analysis and Machine Intelligence (TPAMI)}, 45(7):8954--8968, 2023.

\bibitem{zhou2024test}
Qianyu Zhou, Ke-Yue Zhang, Taiping Yao, Xuequan Lu, Shouhong Ding, and Lizhuang Ma.
\newblock Test-time domain generalization for face anti-spoofing.
\newblock In {\em Proceedings of the IEEE/CVF Conference on Computer Vision and Pattern Recognition (CVPR)}, 2024.

\bibitem{zhou2023instance}
Qianyu Zhou, Ke-Yue Zhang, Taiping Yao, Xuequan Lu, Ran Yi, Shouhong Ding, and Lizhuang Ma.
\newblock Instance-aware domain generalization for face anti-spoofing.
\newblock In {\em Proceedings of the IEEE/CVF Conference on Computer Vision and Pattern Recognition (CVPR)}, pages 20453--20463, 2023.

\bibitem{zou2021geometry}
Longkun Zou, Hui Tang, Ke Chen, and Kui Jia.
\newblock Geometry-aware self-training for unsupervised domain adaptation on object point clouds.
\newblock In {\em Proceedings of the IEEE/CVF International Conference on Computer Vision}, pages 6403--6412, 2021.

\end{thebibliography}
}

\end{document}